# A simulation platform calibration method for automated vehicle evaluation: accurate on both vehicle level and traffic flow level

**Jia Hu** [1], **Junqi Li** [1], **Xuerun Yan** [1], **Jintao Lai** [2], **Lianhua An** [3,*]

[1] Key Laboratory of Road and Traffic Engineering of the Ministry of Education, Tongji University, Shanghai, China

[2] Department of Control Science and Engineering, Tongji University, No.4800 Cao'an Road, Shanghai, China

[3] College of Transport and Communications, Shanghai Maritime University, 1550 Haigang Avenue, Shanghai, 201306, China

**\*** Correspondence: **Lianhua An** (lhan@shmtu.edu.cn)

## Abstract

Simulation test is a fundamental approach for evaluating automated vehicles (AVs). To ensure its reliability, it is crucial to accurately replicate interactions between AVs and background traffic, which necessitates effective calibration. However, existing calibration methods often fall short in achieving this goal. To address this gap, this study introduces a simulation platform calibration method that ensures high accuracy at both the vehicle and traffic flow levels. The method offers several key features:(i) with the capability of calibration for vehicle-to-vehicle interaction; (ii) with accuracy assurance; (iii) with enhanced efficiency; (iv) with pipeline calibration capability. The proposed method is benchmarked against a baseline with no calibration and a state-of-the-art calibration method. Results show that it enhances the accuracy of interaction replication by 83.53% and boosts calibration efficiency by 76.75%. Furthermore, it maintains accuracy across both vehicle-level and traffic flow-level metrics, with an improvement of 51.9%. Notably, the entire calibration process is fully automated, requiring no human intervention.

**Keywords**: simulation test; automated vehicle; calibration; vehicle-to-vehicle interactions

## 1  Introduction

Simulation test is one of the primary approaches for evaluating automated vehicles (AVs). According to a report from Waymo, the global leader in AV technology (The New York Times, 2024), simulation test accounts for 99% of the total miles used in their AV evaluations (Waymo Driver, 2025). Consequently, simulation test is a necessity for AV evaluation.

A realistic replication of interactions between AVs and background traffic is critical for making simulation test reliable. The error in replicating interactions can lead to errors in evaluating emissions, mobility, and driving safety. A 23% replication error may lead to an error in evaluating emissions of up to 15% (G. Song et al., 2013). A 27% replication error may lead to an error in evaluating mobility of up to 31% (Ye & Yamamoto, 2018). A 20% replication error may lead to an error in evaluating driving safety of up to 135% (Sadid & Antoniou, 2024). Therefore, it is crucial to replicate realistic interactions to achieve reliable evaluation.

Calibration is the key to achieving realistic replication of interactions. The interactions are determined by the vehicle models. By tuning parameter values within these vehicle models, the patterns of the interactions are altered (Dai et al., 2022). Since the default parameter values represent only a single pattern of the interactions, which is mostly not the pattern of interest, the default parameter values often fail to replicate realistic interactions. Therefore, these parameter values must be tuned to replicate realistic interactions. This requires calibration—a method specifically designed for parameter tuning.

Existing calibration methods do alter the pattern of the interactions between vehicles. However, their goals are not the realistic replication of interactions. Based on their calibration goals, these calibration methods can be categorized into three groups: macroscopic replication, microscopic replication, and hybrid replication.

Macroscopic replication focuses on replicating traffic flow (Chiappone et al., 2016; Langer et al., 2021; Punzo & Montanino, 2020; Ren et al., 2024). Methods in this group ensure alignment between simulated and real-world traffic flow characteristics. The characteristics, such as traffic volume and average speed, describe the dynamics of the flow by treating all vehicles as a single unit. However, each vehicle has its unique intention and actions, these characteristics cannot accurately describe the action of each vehicle within the flow. Given this, the interactions cannot be replicated accurately. Furthermore, these characteristics are derived from traffic data collected at fixed locations on the road, meaning they only describe vehicle actions at these fixed locations. Since vehicles are constantly moving and taking actions to interact with their surrounding vehicles, interactions are not confined to fixed points but occur continuously along the entire road. Consequently, these characteristics are unable to describe interactions accurately. In summary, macroscopic replication methods are not able to replicate interactions.

Microscopic replication focuses on replicating the behavior of individual vehicles (Kesting & Treiber, 2008; Punzo & Simonelli, 2005; Shikun et al., 2024). Most methods in this group replicate the dynamics of a vehicle reacting to its directly preceding vehicle. Since these methods can calibrate longitudinal actions of the vehicle but fail to calibrate lateral actions, the interactions are unrealistic. Furthermore, although a few methods attempt to calibrate lateral actions by adjusting lane-changing models (Ali et al., 2023; Mohamed et al., 2021), they are fundamentally flawed because they do not fully consider the effects of traffic flow. In the simulated traffic flow of these methods, a vehicle is only affected by its preceding vehicle. However, in the real-world traffic flow, a subject vehicle is also affected by vehicles in adjacent lanes. For instance, if a vehicle in an adjacent lane cuts in front of the subject vehicle, the subject vehicle may need to decelerate to allow the cut-in. If such effects are not well calibrated, such as the frequency and intensity of cut-ins, the actions of the subject vehicle in response to cut-ins would never be authentic. As a result, lacking calibrating traffic flow impairs the authenticity of vehicle actions, leading to unrealistic interactions. Therefore, calibration sorely on the individual vehicle level, without calibrating traffic flow, is not enough for the realistic replication of interactions.

Hybrid replication focuses on replicating both individual vehicle behavior and overall traffic flow (Alhariqi et al., 2022; Hale et al., 2023; U.S. DOT, 2021). These methods replicate the trajectory of each vehicle as it moves within the traffic flow. However, their accuracy has room for improvement. Studies show that the errors of such methods often exceed 60% and can even reach up to 85% (Hale et al., 2023). Since a trajectory describes the actions of a vehicle, this level of accuracy is insufficient to ensure the authenticity of vehicle actions. Consequently, the authenticity of interactions cannot be guaranteed either. Even if the accuracy of these methods was guaranteed, the interactions are still not realistic enough. Since these methods aim to exactly replicate the trajectory of each vehicle at a specific time and location without any stochasticity, vehicle actions after calibration become scripted and deterministic, resulting in a single pattern of interactions (Hale et al., 2023; U.S. DOT, 2021). In contrast, in the real world, the patterns of interactions follow a distribution with inherent variability, even within the same traffic flow. For example, the distance required for a vehicle to complete a lane change can vary significantly, even with the same speed difference in the target lane. As a result, these methods fail to replicate the stochastic nature of interactions in the real world, making the simulation less convincing. Therefore, hybrid replication methods have room for improvement in replicating interactions. In summary, a calibration method capable of replicating realistic interactions is needed.

Even if hybrid replication methods were capable of replicating realistic interactions, they still face significant efficiency problems. These methods replicate interactions by integrating models of AVs and background traffic into the simulation platform. However, since AV models are computationally intensive, their integration into the simulation platform results in a substantial increase in computation time. Studies (Ding et al., 2022; Zhang et al., 2023) show that the computation time for a single AV model in a single simulation step can range from 0.035 seconds to 0.9 seconds. Given that one simulation step typically represents 0.1 seconds of simulated time, this means that simulating 0.1 seconds requires at least 0.035 seconds of computation time when only one AV model is integrated. For a simulation lasting 1 hour (36,000 simulation steps), the computation time amounts to at least 21 minutes. This increases further as more AVs are involved in the simulation. When considering the large number of simulations required by these methods—typically ranging from 1,000 to 10,000 simulations (Chao Zhang et al., 2017; L. Li et al., 2016; Pourabdollah et al., 2017; Tang et al., 2024)—the total computation time becomes prohibitively long. Taking the minimum number of 1,000 simulations as an example, with each simulation running for 21 minutes, the total computation time would span approximately 15 days. As a result, these methods are highly time-consuming. Moreover, the calibration process can be further prolonged due to the need for human intervention. Human intervention is primarily required to identify the choices of calibration parameters. Based on experience, humans select and adjust parameters, run simulations, and evaluate the results of simulations to determine whether the parameters shall be considered for the calibration (Han et al., 2013). However, since human work is limited to a few hours per day (e.g. 8 hours), interventions cannot always occur timely, often pausing the calibration process. This inevitably extends the calibration time, further reducing efficiency. In summary, existing calibration methods require enhancement to improve their calibration efficiency.

Given the shortcomings of existing calibration methods, this research proposes a simulation platform calibration method for AV evaluation, which is accurate on vehicle level and traffic flow level. It bears the following features:

- With the capability of calibration for vehicle-to-vehicle interaction.
- With accuracy assurance.
- With enhanced efficiency.
- With pipeline calibration capability.

## 2 Highlights

This section provides the rationale for how the features of this study are achieved.

- **With the capability of calibration for vehicle-to-vehicle interaction:** The attributes affecting vehicle-to-vehicle interactions are identified and integrated into the calibration objectives. Vehicle-to-vehicle interactions include lateral and longitudinal interactions. The proposed method decomposed these interactions into specific attributes. The attributes of longitudinal interactions, such as car-following headway, the speed of the following vehicle, describes how the behavior of a following vehicle is affected by the movement of the leading vehicle. The attributes of lateral interactions, such as lane-changing distance, describe the influence of lateral movement of vehicles caused by surrounding vehicles in adjacent lanes. These attributes are incorporated into calibration objectives, guiding the calibration process to generate vehicle-to-vehicle interactions that closely reflect real-world interactions.
- **With accuracy assurance:** Accuracy is assured by the additional consideration of replicating vehicle-to-vehicle interactions. This is achieved by calibrating the maneuver patterns of vehicles during interactions. To capture the stochastic nature of individual vehicle behaviors, the distribution of maneuver patterns for each vehicle is included in the calibration objectives, guiding the process to accurately replicate the driving behavior of individual vehicles in response to their surrounding vehicles. Additionally, to capture a link between individual vehicles in the interactions and the overall traffic flow, the collective distribution of maneuver patterns across all vehicles during vehicle-to-vehicle interactions is integrated into the calibration objectives. This guides the calibration process to precisely replicate the characteristics of the traffic flow as a whole. Hence, accuracy is assured in replicating both individual vehicle behaviors and traffic flow.
- **With enhanced efficiency:** Efficiency is enhanced by reducing the number of test cases, which is achieved through two strategies: i) Calibration decomposition: Unlike traditional hybrid methods that replicate both individual vehicle behavior and traffic flow simultaneously in a single calibration problem, this study decomposes the problem into two subproblems: one for individual vehicle behaviors and one for traffic flow. This reduces the solution space significantly, making the number of test cases required decrease for achieving the same level of calibration accuracy. ii) Only cases that are critical are tested: To identify parameters for calibration, this study tests only critical cases instead of all possible cases. A case selection approach is proposed to select these critical cases. The approach ensures that each parameter's effect on accuracy is quantified independently. This avoids the cases where parameter effects influence each other, which constitute the majority of all cases. Since these cases do not contribute to quantifying each parameter's effect, removing them does not hinder the identification of calibration parameters but significantly improves efficiency.
- **With pipeline calibration capability:** This study achieves pipeline calibration capability by automating the entire calibration process. The process involves selecting calibration parameters, generating parameter value pairs, running simulations, and evaluating results. As mentioned in the Introduction, existing methods often require human intervention in selecting calibration parameters, making the process not fully automated. To address this, this study proposes an automated approach to identify parameters for calibration. This approach analyzes how changes in the value of each parameter impact calibration accuracy and selects the parameters with the greatest impact for calibration. With this approach, the study achieves fully automated calibration, enabling pipeline calibration capability.

# 3 Problem statement

This study aims to develop a realistic simulation platform for AV evaluation. Figure 1 illustrates a typical scenario: an AV travels from its origin to its destination, encountering various background traffic along the way. For clarity, in this study, AVs are referred to as subject vehicles. Background traffic is classified into surrounding vehicles and more distant vehicles. Surrounding vehicles are those within the detection range of the subject vehicle and directly interact with it, whereas more distant vehicles lie outside this range and have minimal direct interaction. As all vehicles move, their relative positions change dynamically, causing vehicles to transition between the surrounding and more distant categories. Detailed information on both the AV and its surrounding vehicles is obtained from the AV's detection data.

The simulation platform is calibrated using this detection data to achieve three objectives: (i) accurate replication of the subject vehicle's behavior, as it serves as the reference; (ii) accurate replication of interactions between the subject vehicle and surrounding vehicles; and (iii) accurate replication of traffic flow characteristics. These objectives are achieved by calibrating the vehicle models of both surrounding and more distant vehicles.

Notably, although Figure 1 depicts a single AV for clarity, the proposed method supports the integration of multiple AVs and accommodates a variety of AV models.

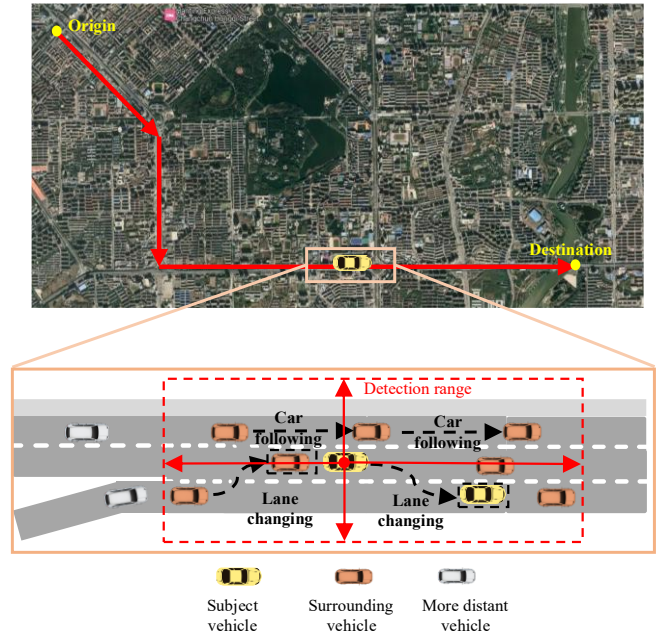

**Figure 1** An example of a typical scenario for calibration.

# 4 Methodology

## 4.1 System framework

The system framework of the proposed method is illustrated in Figure 2. The framework consists of the following modules:

- **Measures of performance (MoPs) selection:** Selects MoPs to guide the calibration process. These MoPs include vehicle-level MoPs and traffic-level MoPs. Details are provided in the following.
- **Field data processing:** Processes field data to extract the real-world values of the selected MoPs. The work of this module includes data preprocessing (e.g., filtering and smoothing), identification of driving behaviors (e.g., car-following and lane-changing), and calculation of the real-world MoP values.
- **Simulation network construction:** Builds a simulation network that replicates the real-world road environment traveled by the subject vehicle.
- **Pipeline calibration:** Performs a fully automated calibration process to replicate both traffic flow and individual vehicle behaviors. The module adopts a two-stage architecture: Stage 1 focuses on accurately replicating traffic flow, while Stage 2 focuses on accurately replicating individual vehicle behaviors. Further details are provided in the following.

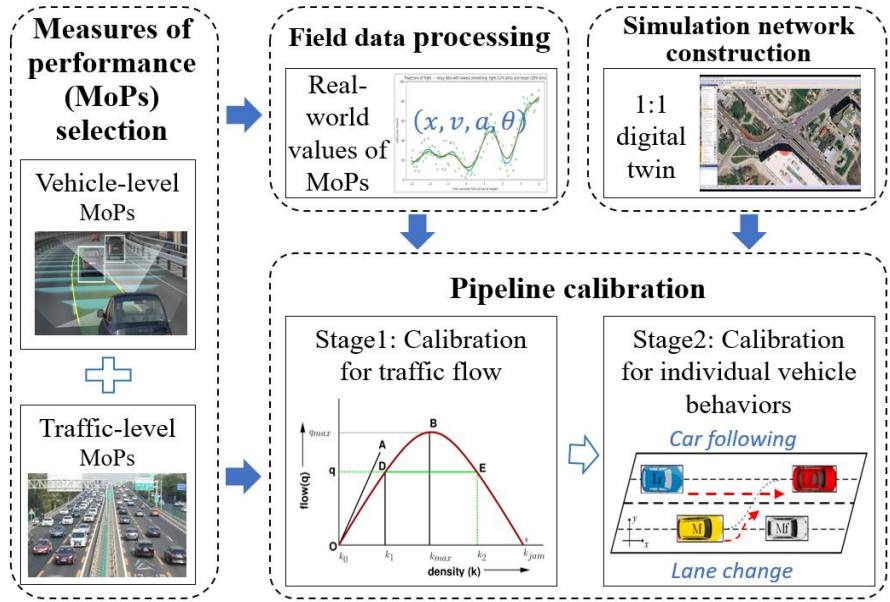

**Figure 2** The framework of the proposed method.

## 4.2 MoPs selection

To ensure the calibration method accurately replicates both individual vehicle behaviors and traffic flow, the selected MoPs include vehicle-level MoPs and traffic-level MoPs. Details are as follows.

### 4.2.1 Vehicle-level MoPs

Vehicle-level MoPs capture attributes affecting vehicle-to-vehicle interactions. They are categorized based on the following four factors:

*Interaction behavior:* The type of vehicle behavior involved in the interaction, categorized into lateral (lane changes, overtaking, cut-ins) and longitudinal (car-following) behaviors.

*Studied object:* The kinds of vehicles involved in the interaction. A distinction is made between the subject vehicle and surrounding vehicles, as their behaviors differ and thus require separate consideration when defining MoPs.

*Statistic quantity:* Metrics that describe the statistical characteristics of interactions, such as averages, distributions, and quantiles.

*Maneuver features:* Key features of a specific interaction maneuver. For instance, in a lane change, key features include the lateral distance required and the time taken to complete the lane change maneuver.

By combining these factors in different ways, various vehicle-level MoPs can be formulated. A detailed list of suggested vehicle-level MoPs is provided in Table 2 (Appendix A).

### 4.2.2 Traffic-level MoPs

Traffic-level MoPs capture attributes that influence traffic flow dynamics encountered by the subject vehicle. These MoPs characterize both the subject vehicle and its surrounding traffic, ensuring that the simulated traffic flow aligns with real-world traffic flow.

***Subject vehicle characteristics***: These MoPs describe the travel time of the subject vehicle, ensuring consistency between the timing of traffic flow data collection in simulation and the real world.

***Background traffic characteristics***: These MoPs characterize traffic flow dynamics by considering the subject vehicle and interactive vehicles as a single unit, capturing the overall traffic dynamics.

A detailed list of suggested traffic-level MoPs is provided in Table 3 (Appendix A).

## 4.3 Pipeline calibration

### 4.3.1 Stage 1: Calibration for traffic flow

**Problem formulation**

Stage 1 aims to replicate accurate traffic flow. The calibration can be formulated as an optimization problem, defined by its decision variables, objective function, and constraints as follows:

*Decision variables:* The decision variables are the calibration parameters of traffic flow, as defined below:

$$x^{s1} = [x_1^{s1}, x_2^{s1}, \cdots, x_{M^{s1}}^{s1}] \tag{1}$$

where $x^{s1}$ is the parameter vector to be calibrated. $x_j^{s1}$ represents the $j^{th}$ parameter to be calibrated, specifically the vehicle input at entrance roads in this stage. $M^{s1}$ is the total number of vehicle inputs.

*Objective function:* The objective is to minimize the discrepancy between the simulated and real-world values of traffic-level MoPs. The objective function is formulated as follows:

$$\min f^{s1}(y^{s1,field}, y^{s1,sim}) = \sum_{i=1}^{N^{s1}} \frac{|y_i^{s1,field} - y_i^{s1,sim}|}{y_i^{s1,field}} \tag{2}$$

$$y^{s1,sim} = F^{s1}(x^{s1}) \tag{3}$$

where $f^{s1}(\cdot)$ represents the goodness-of-fit function in Stage 1. $y^{s1,field}$ represents the real-world MoP values obtained from field data at Stage 1. In this study, the selected traffic-level MoPs are the average speed of the subject vehicle and the average density across all traveled roads of the subject vehicle. $y^{s1,sim}$ denotes the corresponding simulated MoP value at Stage 1. $N^{s1}$ is the number of selected MoPs. $i$ is the index of a specific MoP. $F^{s1}(\cdot)$ is the mapping function, which describes the traffic flow-level relationship between calibration parameters and the resulting simulated MoP values.

*Constraints:* Each calibration parameter has its boundaries, as defined follows:

$$\underline{x}_j^{s1} \leq x_j^{s1} \leq \overline{x}_j^{s1}, \forall x_j^{s1} \in X^{s1} \tag{4}$$

$$\underline{x}_j^{s1} = (1-\delta)x_{j,0}, \forall x_j^{s1} \in X^{s1} \tag{5}$$

$$\overline{x}_j^{s1} = (1+\delta)x_{j,0}, \forall x_j^{s1} \in X^{s1} \tag{6}$$

where $\underline{x}_j^{s1}$ and $\overline{x}_j^{s1}$ are the lower and upper bounds of the $j^{th}$ parameter to be calibrated, respectively. $x_{j,0}$ is the initial value derived from the real-world data. $\delta$ is a scaling factor that defines the range of allowable parameter variation.

**Solving algorithm**

To find a set of optimal calibration parameter values that enable the accurate replication of traffic flow, a solving algorithm is proposed. The algorithm, named Simulation-based Orthogonal Array Testing (SOTA), integrates Orthogonal Array Testing (OAT) with a simulation-in-the-loop architecture. In this algorithm, OAT efficiently reduces the number of cases while maintaining comprehensive coverage of all possible cases (Yu et al., 2011). Simulations are to evaluate the performance of each case. The procedure consists of the following steps:

i) Determination of calibration parameter values

Let $L$ be the set of levels of associated with each parameter, defined as:
$$L = \{l_1, l_2, l_j \dots, l_N\} \tag{7}$$
where $l_j$ is the total number of levels for the $j^{th}$ parameter. $N$ is the number of calibration parameters. In this stage, $N = M^{s1}$.

For any parameter, its possible values are determined based on its predefined range and level. Assuming uniform sampling across the range, the possible values are given as follows:

$$V_j = \left\{ \underline{x}_j^{s1} + k \frac{\overline{x}_j^{s1} - \underline{x}_j^{s1}}{l_j - 1} \mid k = 0,1,2, \dots, l_j - 1 \right\} \tag{8}$$

where $V_j$ is the set of possible values of the $j^{th}$ parameter to be calibrated. $k$ is the index of the level.

ii) Test case design

The test cases are designed based on orthogonal arrays. The orthogonal arrays are designed to refer to the handbook of statistical methods (Guthrie, 2020). Each level in the orthogonal array corresponds to a specific value of calibration parameters within the predefined range, forming a complete set of test cases. For example, consider three vehicle inputs (inflows) from Road A, Road B, and Road C, each with two levels. The designed orthogonal array is shown in Figure 3(a). Assuming the initial vehicle inputs are 800 *pcu/h*, 1000 *pcu/h*, and 1200 *pcu/h* (obtained from the real-world data), their value ranges are set as [640 *pcu/h*, 960 *pcu/h*], [800 *pcu/h*, 1200 *pcu/h*], and [960 *pcu/h*, 1440 *pcu/h*]. By mapping each level in the orthogonal array to its value within these ranges, the resulting set of test cases is shown in Figure 3(b).

| Experiment No | $x_1^{s1}$ | $x_2^{s1}$ | $x_3^{s1}$ |
|---|---|---|---|
| 1 | level 1 | level 1 | level 1 |
| 2 | level 1 | level 2 | level 2 |
| 3 | level 2 | level 1 | level 2 |
| 4 | level 2 | level 2 | level 1 |

| Case No | $q_A$ (pcu/h) | $q_B$ (pcu/h) | $q_C$ (pcu/h) |
|---|---|---|---|
| 1 | 640 | 800 | 960 |
| 2 | 640 | 1200 | 1440 |
| 3 | 960 | 800 | 1400 |
| 4 | 960 | 1200 | 960 |

(a) The orthogonal array.  (b) Test cases.

**Figure 3** An example of test case design.

iii) Case evaluation by simulation

Each test case is evaluated as follows:

$$a(c) = 1 - f^{s1}(y^{s1,field}, y^{s1,sim}) \quad (9)$$

where $a(c)$ is the calibration accuracy resulting from case $c$.

The detailed solving process for Stage 1 is presented in **Algorithm 1.**

---

**Algorithm 1** Simulation-based orthogonal array testing algorithm (SOAT)

---

**Input:** Calibration parameters $x^{s1}$, parameter levels $L = \{l_1, l_2, ..., l_N\}$, the accuracy threshold $a_{thre}$

**Output:** The optimal set of parameter values $c_{best}$, the maximum accuracy $a_{max}$

1: Initialize $c_{best} = None$, $a_{max} = 0$
2: //Step 1. Determine values of each calibration parameter $V_j$
3: **for** $j$ in (1, $N$) **do**
4:     Calculate $V_j$ according to Equation (8)
5: **end for**
6: Generate test cases //Step2. Design cases based on an orthogonal array
7: **for** each case $c$ **do** //Step 3. Evaluate each test case in simulation
8:     Configure the simulation platform parameters as defined by each test case
9:     Run simulation and evaluate the simulation results $y^{s1,sim}(c)$
10:     Evaluate accuracy $a(c)$ according to Equation (9)
11:     **if** $a(c) > a_{max}$ **then**
12:         $a_{max} \leftarrow a(c)$
13:         $c_{best} \leftarrow c$
14:     **end if**
15:     **if** $a(c) >= a_{thre}$ **then**
16:         break
17:     **end if**
18: **end for**
19: Set the simulation platform with $c_{best}$
20: **return** $c_{best}$ and $a_{max}$

---

### 4.3.2 Stage 2: Calibration for individual vehicle behaviors

Stage 2 focuses on replicating individual vehicle behaviors. The calibration framework is shown in Figure 4. It includes two phases: Phase I explores the parameter space spanned by all available parameters of vehicle behavior. It combines SOAT with Range Analysis (RA) to identify calibration parameters that significantly impact calibration accuracy. Phase II optimizes the values of the identified calibration parameters. It employs a Simulation-based Adaptive Genetic Algorithm (SAGA) to determine the optimal parameter values, ensuring a more accurate replication of individual vehicle behaviors.

Notably, vehicle behavior parameters are derived from the vehicle models used in simulations. In this study, multiple models are adopted: the subject vehicle model and surrounding vehicle model adopt an integrated behavioral and motion planner developed by (Zhang et al., 2023), while more distance vehicle model utilizes VISSIM's default W99 car-following model (Durrani et al., 2016) and lane-changing models (Al-Msari et al., 2024). Their parameters are listed in Table 4 (Appendix B). Since the proposed method is applicable to various vehicle models, other commonly used vehicle models and their suggested parameters are also provided in the table.

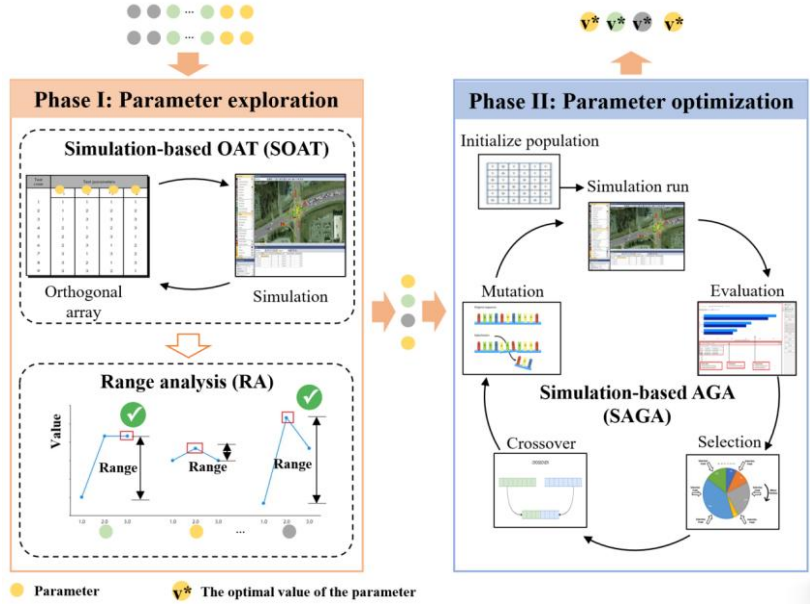

**Figure 4** The process of Stage 2.

**Phase I: Parameter exploration**

### Problem formulation

The problem of Phase I in Stage 2 is formulated as an optimization problem. Its decision variables, objective function, and constraints are outlined as follows:

*Decision variables:* The decision variables are represented by all available vehicle behavior parameters. They are defined by:

$$x^{s2} = [x_1^{s2}, x_2^{s2}, \cdots, x_{M^{s2}}^{s2}] \tag{10}$$

where $x^{s2}$ is the parameter vector to be calibrated at stage 2. $x_j^{s2}$ represents the $j^{th}$ parameter to be calibrated at stage 2. $M^{s2}$ is the total number of all available parameters relating to individual vehicle behaviors.

*Objective function:* The objective is to minimize the differences of individual vehicle behaviors between the selected vehicle-level MoPs in the simulation and the real world. The objective function is defined as:

$$\min f^{s2}(y^{s2,field}, y^{s2,sim}) = \sum_{i=1}^{N^{s2}} \frac{|y_i^{s2,field} - y_i^{s2,sim}|}{y_i^{s2,field}} \quad (11)$$

$$y^{s2,sim} = F^{s2,I}(x^{s2}) \quad (12)$$

where $f^{s2}(\cdot)$ is the goodness-of-fit function in stage 2. $y^{s2,field}$ represents the real-world values of vehicle-level MoPs. In this study, the selected vehicle-level MoPs include the average car-following headway of surrounding vehicles, the average lane-changing distance of the subject vehicle, the average lane-changing distance of surrounding vehicles, and the number of cut-ins for the subject vehicle. $y^{s2,sim}$ represents the simulated values of vehicle-level MoPs. $N^{s2}$ is the number of selected MoPs in stage 2. $i$ denotes the $i^{th}$ MoP. $F^{s2,I}(\cdot)$ is the mapping function, which describes the vehicle-level relationship between calibration parameters and the resulting simulated MoP values.

*Constraints:* Each calibration parameter has its boundaries, defined as:

$$\underline{x}_j^{s2} \leq x_j^{s2,I} \leq \overline{x}_j^{s2}, \forall j = 1,2\ldots, M^{s2} \quad (13)$$

where $\underline{x}_j^{s2}$ and $\overline{x}_j^{s2}$ are the lower and upper boundary of the $j^{th}$ parameter to be calibrated in stage 2, respectively.

**Solving algorithm**

The solving algorithm integrates SOAT with Range Analysis (RA) to systematically identify the calibration parameters. SOAT selects cases that are critical in the entire parameter space and then evaluates them. RA identifies the parameters that significantly impact accuracy based on the evaluation results of SOAT (H. X. Li et al., 2020). SOAT algorithm has been presented in Algorithm 1. The key procedures of RA adopted are presented as follows.

i) Accuracy evaluation for each parameter level

The calibration accuracy is computed for each parameter level by grouping accuracy values across cases. For each parameter $x_j^{s2}$, the average accuracy at each level $l$ is calculated as:

$$\bar{a}(x_j^{s2}, l) = \frac{1}{N_{l,j}} \sum_{z=1}^{N_{l,j}} a_{j_z,l} \quad (14)$$

where $N_{l,j}$ is the total number of cases that the $j^{th}$ parameter is at level $l$. $a_{j_z,l}$ is the calibration accuracy of the $z^{th}$ case that the $j^{th}$ parameter is at level $l$.

ii) Range analysis for each parameter

The range measures how much the accuracy varies when the parameter changes across its levels. For each parameter, the range is calculated by:

$$R_j = \max_{l \in \{1,2,\ldots,l_j\}} \bar{a}(x_j^{s2}, l) - \min_{l \in \{1,2,\ldots,l_j\}} \bar{a}(x_j^{s2}, l) \quad (15)$$

where $R_j$ is the accuracy range of the $j^{th}$ parameter.

iii) Selection of critical parameters

Critical parameters are those parameters that have the greatest impact on the accuracy. They are defined as the ones with the largest accuracy ranges. Details are as follows:

Let $\pi$ be a permutation of the index set of calibration parameters that arranges them in descending order according to their corresponding accuracy ranges. The resulting sequence of

permuted indices $(\pi(1), \pi(2), \ldots, \pi(M^{s2}))$ satisfies the following condition:

$$R_{\pi(1)} \geq R_{\pi(2)} \geq \cdots \geq R_{\pi(M^{s2})} \tag{16}$$

The set of critical parameters is constructed by selecting the parameters whose indices are the first $K$ elements in this permuted sequence, defined as:

$$C_K = \{x^{s2}_{\pi(j)} | j = 1,2, \ldots, K\} \tag{17}$$

where $C_K$ is the set of critical parameters.

iv) Identification of the optimal parameter combination

For each parameter, identify the optimal level that yields the highest average accuracy $\bar{a}(x^{s2}_j, l)$. By combining the optimal levels of all parameters, the best parameter combination $c_{best}$ is obtained.

Detailed solving process based on RA is presented in **Algorithm 2.**

---
**Algorithm 2** Range analysis (RA)
---
**Input:** The designed orthogonal array, the accuracy of each test case $a(c)$, the number of critical parameters to be calibrated $K$

**Output:** The set of critical calibration parameters $C_K$, the best parameter combination $c_{best}$

1:     Initialize ranges of all parameters $R = \{\}$, critical parameters $C_K = [\,]$, average accuracy $\bar{a} = \{\}$
2:     **for** $x^{s2}_j$ in $x^{s2}$ **do** //step 1. Calculate the average accuracy
3:        Initialize level averages: $\bar{a}(x^{s2}_j) = \{\}$
4:        **for** each level $l \in \{1,2,\ldots, l_j\}$ **do**
5:           Extract all cases $c(x^{s2}_j, l)$ from the designed orthogonal array
6:           Compute $\bar{a}(x^{s2}_j, l)$ according to Equation (14)
7:           Store the average accuracy: $\bar{a}(x^{s2}_j)[l] = \bar{a}(x^{s2}_j, l)$
8:        **end for**
9:        Compute the range $R_j$ according to Equation (15) //step 2. Calculate the range
10:    Store $R_j$ in $R$: $R(x^{s2}_j) = R_j$
11:    **end for**
12:    Find the set of critical parameters that with top-$K$ largest accuracy ranges $C_K$ //step3: Select critical parameters for calibration
13:    Find $c_{best}$ //step4: Identify the optimal level combination
14:    Output $C_K$, $c_{best}$

---

## Phase II: Parameter optimization

### Problem formulation

The problem formulation in Phase II is similar to Phase I. The main difference lies in decision variables. Instead of considering all available parameters, Phase II focuses only on the critical calibration parameters identified in Phase I, defined as:

$$x^{s2,II} = [x^{s2}_{\pi(1)}, x^{s2}_{\pi(2)}, x^{s2}_{\pi(j)} \cdots x^{s2}_{\pi(K)}] \tag{18}$$

where $x^{s2,II}$ is the parameter vector to be calibrate in phase II at stage 2.

**Solving algorithm**

Phase II proposed the Simulation-based Adaptive Genetic Algorithm (SAGA) to optimize the values of the critical calibration parameters. SAGA integrates the Adaptive Genetic Algorithm (AGA) (Wu et al., 1998), within the simulation-in-the-loop framework. AGA employs an iterative process to find an optimal parameter combination—a set of parameter values that yields the best performance according to the accuracy. Each iteration has a population of parameter combinations which are generated by three operators: selection, crossover, and mutation. During the iterative process, AGA adaptively adjusts the probability of crossover ($P_c$) and mutation operator ($P_m$) based on the calibration accuracy of parameter combinations in the current iteration. This enhances both the convergence speed and calibration accuracy. The key details of the three operators are as follows:

i) Selection operator

The selection operator constructs a mating pool that serves as the foundation for identifying high-accuracy calibration parameters. This mating pool consists of parameter combinations selected from the current population, based on a selection probability. The selection probability is designed to favor parameter combinations with higher calibration accuracy, and is defined as follows:

$$Prob(p_i) = f(p_i) / \sum_{j=1}^{N_p} f(p_j) \tag{19}$$

$$f(p_i) = 1 - f^{s2}(y^{s2,field}, y^{s2,sim}) \tag{20}$$

where $p_j$ is the $j^{th}$ parameter combination in the current population. A parameter combination refers to a set of assigned values corresponding to each component of the parameter vector $x^{s2,II}$. $Prob(p_i)$ is the selection probability of $p_i$. $f(p_i)$ is the calibration accuracy of $p_i$. $N_p$ is the number of parameter combinations.

ii) Crossover operator

The crossover operator selects two parameter combinations from the mating pool (referred to as a pair of parents ($p_a, p_b$)) to generate two new combinations (referred to as offspring($c_1, c_2$)). For each parent pair, crossover occurs with a certain probability, which is dynamically adjusted based on their calibration accuracy. The crossover probability is defined as follows:

$$P_c = \begin{cases} P_{c,max} - \dfrac{(P_{c,max} - P_{c,min})}{f_{max,cur} - f_{avg,cur}}(f - f_{avg,cur}), f > f_{avg,cur} \\ P_{c,max}, f \leq f_{avg,cur} \end{cases} \tag{21}$$

where $P_{c,max}$ and $P_{c,min}$ are the predefined maximum value and the minimum value of crossover probability. $f_{max,cur}$ is the maximum accuracy in the current generation. $f_{avg,cur}$ is the average accuracy of the current generation. $f$ is the average accuracy of the parent pair.

iii) Mutation operator

The mutation operator introduces random variations into the parameter combination. With a specified mutation probability, it modifies the values of two parameters within the combination, defined as follows:

$$P_m = \begin{cases} P_{m,max} - \dfrac{(P_{m,max} - P_{m,min})}{f_{max,cur} - f_{avg,cur}}(f - f_{avg,cur}), f > f_{avg,cur} \\ P_{m,max}, f \leq f_{avg,cur} \end{cases} \tag{22}$$

where $P_{m,max}$ and $P_{m,min}$ are the predefined maximum and the minimum value of the mutation probability, respectively.

The detailed solving process is presented in **Algorithm 3**.

**Algorithm 3** Simulation-based adaptive genetic algorithm (SAGA)

**Input:** population size $N_p$, maximum generations $G_{max}$, crossover probability boundaries ($P_{c,min}$, $P_{c,max}$), mutation probability boundaries ($P_{m,min}$, $P_{m,max}$), the accuracy threshold $a_{thre}$

**Output:** The optimal accuracy $f_{max}$, the best parameter value combination $p_{best}$

1:     Initialize $iter = 0$, the first population $P_{iter} = \{p_j | j = 1, 2 \cdots N_p\}$, $f_{max} = -inf$, $p_{best} = None$ #Step 1. Initialization
2:     **while** $iter <= G_{max}$ **do**
3:        Configure the simulation platform parameters based on parameter combinations in the current population
4:        Run simulations and evaluate accuracy of these parameter combinations $F^{s2,I}(p_i), \forall p_i \in P_{iter}$ # Step2: Accuracy evaluation
5:        The optimal accuracy at current population: $f_{max,cur} = \max_i f(p_i)$
6:        The best combination at current population: $p_{best,cur} = p_{iter}\left(\operatorname*{argmax}_i f(p_i)\right)$
7:        **if** $f_{max,cur} > f_{max}$ **then** #update best solution
8:            $f_{max} \leftarrow f_{max,cur}$, $p_{best} \leftarrow p_{best,cur}$
9:        **end if**
10:      **if** $f_{max} > a_{thre}$ **then** # check stopping criteria
11:         break
12:      **end if**
13:      Construct mating pool $M = Selection(p_{iter})$ //Step3: selection operator
14:      Initialize $offspring = []$
15:      **for** $i$ in 0 to $\frac{len(M)}{2}$ **do:**
16:         Select a pair of parents: $(p_a, p_b) = (M[i], M[i + 1])$
17:         $f = \frac{[f(p_a)+f(p_b)]}{2}$, $f_{max} = \max(f(p_a), f(p_b))$
18:         Update $P_c$ and $P_m$ according to Equation (21) – (22)
19:         **if** random number $< P_c$ **then** # step4: crossover operator
20:            $(c_1, c_2) = crossover(p_a, p_b)$
21:         **else**
22:            $(c_1, c_2) = crossover(p_a, p_b)$
23:         **if** random number $< P_m$ **then** # step5: mutation operator
24:            $c_1 = mutation(c_1)$
25:         **if** random number $< P_m$ **then**
26:            $c_2 = mutation(c_2)$
27:         Extend $(c_1, c_2)$ into $offspring$
28:      **end for**
29:      $iter = iter + 1$
30:      $p_{iter} = offspring[:N_p]$

31: **end while**

32: **Return** $f_{max}, p_{best}$

## 5 Evaluation

In this section, the proposed method is evaluated. The evaluation is based on the following aspects: (i) calibration capability of vehicle-to-vehicle interaction, (ii) calibration capability of traffic flow, and (iii) calibration efficiency.

### 5.1 Simulation platform

VISSIM is selected for evaluation. It ran on a computer with an Intel i5-13500H 2.60 GHz processor and 16GB RAM.

### 5.2 Data preparation

The data was collected by an AV equipped with various sensors and cameras. It includes: ***i) Environmental data:*** Provides information about the traffic environment in which the subject vehicle travels. ***ii) Subject vehicle data:*** Records vehicle states of the AV itself. ***iii) Surrounding vehicle data:*** Captures the states of vehicles detected by the subject vehicle. The data was collected between 14:15 and 14:45 on June 8, 2023, in Changchun, Jilin Province, China.

A detailed description of the data is provided in **Table 1**.

Table 1 Description of the data.

| Category | Variable name | Description | Unit |
|---|---|---|---|
| Environmental data | Timestamp | Time of data recording | / |
|  | Road name | Name of the road segment where data was collected. | / |
|  | Speed limit | Legal speed limit of the current road segment. | $km/h$ |
| Subject vehicle data | Speed | Instantaneous speed of the subject vehicle | $km/h$ |
|  | Yaw rate | Angular velocity of the subject vehicle around the vertical axis | $rad/s$ |
|  | Longitude | Longitude of the subject vehicle | $rad$ |
|  | Latitude | Latitude of the subject vehicle | $rad$ |
|  | Acceleration | Longitudinal acceleration (acceleration: +, deceleration: -). | $m/s^2$ |
|  | Lane ID | Identifier of the current lane, where 1 is the rightmost lane when viewed in the driving direction, increasing from right to left. | / |
|  | Lane distance | Lateral distance from vehicle centerline to nearest | $m$ |

| | | lane marking (left: +, right: -). | |
|---|---|---|---|
| **Surrounding vehicle data** | Vehicle ID | Unique identifier assigned to each detected surrounding vehicle. | / |
| | Surrounding vehicle's Lane ID | Lane identifier of the surrounding vehicle (same numbering rule as subject vehicle). | / |
| | Relative longitudinal position | Longitudinal distance from the subject vehicle to the surrounding vehicle (front: +, rear: -). | $m$ |
| | Relative lateral position | Lateral distance from the subject vehicle to the surrounding vehicle (left: +, right: -). | $m$ |
| | Absolute velocity | Speed of the surrounding vehicle in the global coordinate system. | $km/h$ |
| | Vehicle heading | Orientation angle relative to road direction (0°: parallel, ±180°: opposing direction). | ° |

## 5.3 Experiment design

### 5.3.1 Testbed

Figure 5 illustrates the testbed used in this study. Figure 5(a) shows the real-world route taken by the subject vehicle, as recorded in the collected data. This route includes sections of the Xibu Expressway and Nanbu Expressway in Changchun, covering a total distance of 10.15 km. Based on this route, a corresponding simulation network was constructed, as shown in Figure 5(b).

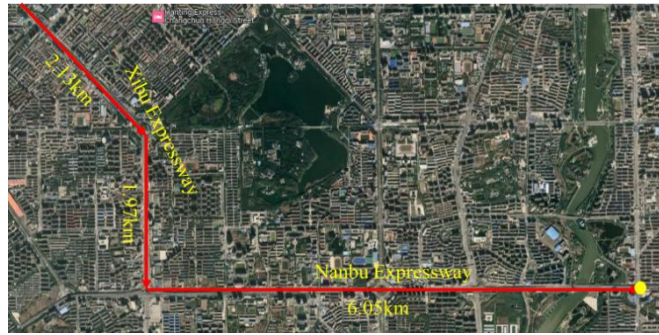

(a) The satellite view of the network.

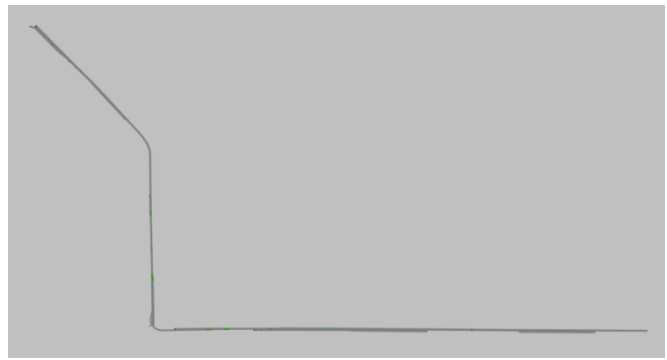

(b) The simulation network.

**Figure 5** Testbed.

### 5.3.2 Tested methods

The tested methods are as follows:
• **No calibration:** This method does not conduct any calibration. Traffic flow parameters remain at their initial values and vehicle behavior parameters are set to their default values. The method serves as a common benchmark.
• **State-of-the-art (SOTA) method:** This method falls into the category of macroscopic replication (Paz et al., 2015). This method focuses on achieving traffic flow-level accuracy and does not consider vehicle-level accuracy. In addition, it does not propose designs to enhance calibration efficiency.
• **The proposed method:** It is a hybrid replication method that focuses on both traffic flow-level accuracy and vehicle-level accuracy. It employs the proposed efficiency enhancement design.

### 5.3.3 Measures of effectiveness (MoEs)

The performance of the proposed method is evaluated across three aspects: vehicle-level accuracy, traffic flow-level accuracy, and calibration efficiency.

**Vehicle-level accuracy**

Vehicle-level accuracy measures the calibration capability of vehicle-to-vehicle interactions, which involves both longitudinal and lateral aspects. It is quantified as the total error across all adopted MoEs for these two aspects, defined by:

$$e_{veh} = \sum_{m \in Lon} e_m + \sum_{n \in Lat} e_n \quad (23)$$

where $e_m$ and $e_n$ represent the values of MoE $m$ and $n$, respectively. $Lon$ and $Lat$ represents the set of MoEs used to evaluate longitudinal accuracy and lateral accuracy, respectively.

The adopted MoEs for accuracy of longitudinal and lateral interactions are detailed below:

1) *Longitudinal accuracy:* Two MoEs are adopted to evaluate the calibration capability of longitudinal interactions:

(i) Error in subject vehicle's headway: The relative error between the simulated and real-world average time headway of the subject vehicle, as defined by:

$$e_{t,sub} = \left| \frac{\sum_{t=1}^{N_{sub}^{sim}} t_{sub,t}^{sim}}{N_{sub}^{sim}} - \frac{\sum_{t=1}^{N_{sub}^{real}} t_{sub,t}^{real}}{N_{sub}^{real}} \right| / \frac{\sum_{t=1}^{N_{sub}^{real}} t_{sub,t}^{real}}{N_{sub}^{real}} \times 100\% \quad (24)$$

where $t_{sub,t}^{sim}$, $t_{sub,t}^{real}$ are the *t*-th time headway of the subject vehicle in simulation data and real-world data, respectively. $N_{sub}^{sim}$ and $N_{sub}^{real}$ are the total number of headways observations in simulation and real-world data.

(ii) Error in background vehicles' headway: The relative error between simulated and real-world average time headway of background vehicles. It is calculated based on the sampled data from surrounding vehicles observed by the subject vehicle. The MoE is defined by the following equations:

$$t_{bg}^{sim} = \frac{1}{N_{surr}^{sim}} \sum_{k=1}^{N_{surr}^{sim}} \frac{\sum_{t=1}^{N_k^{sim}} t_{k,t}^{sim}}{N_k^{sim}} \qquad (25)$$

$$t_{bg}^{real} = \frac{1}{N_{surr}^{real}} \sum_{k=1}^{N_{surr}^{real}} \frac{\sum_{t=1}^{N_k^{real}} t_{k,t}^{real}}{N_k^{real}} \qquad (26)$$

$$e_{t,bg} = \frac{|t_{bg}^{sim} - t_{bg}^{real}|}{t_{bg}^{real}} \times 100\% \qquad (27)$$

where $t_{bg}^{sim}$ is the average time headway of background vehicles in the simulation, $t_{k,t}^{sim}$ denotes the $t$-th headway of the $k$-th surrounding vehicle, $N_k^{sim}$ is the number of observations for that vehicle, $N_{surr}^{sim}$ is the total number of surrounding vehicles observed, $t_{bg}^{real}$ is the corresponding value from the real-world data using analogous notation, and $e_{t,bg}$ is the relative error.

2) *Lateral accuracy:* Three MoEs are adopted to evaluate the calibration capability of lateral interactions:

(i) Error in subject vehicle's lane-changing distance: The relative error between the simulated and real-world average lane-changing distance of the subject vehicle. The lane-changing distance refers to the gap between the leading and following vehicles in the target lane at the moment when the subject vehicle initiates a lane change. The MoE is defined as follows:

$$e_{d,sub} = \left| \frac{\sum_{i=1}^{L_{sub}^{sim}} d_{sub,i}^{sim}}{L_{sub}^{sim}} - \frac{\sum_{i=1}^{L_{sub}^{real}} d_{sub,i}^{real}}{L_{sub}^{real}} \right| / \frac{\sum_{i=1}^{L_{sub}^{real}} d_{sub,i}^{real}}{L_{sub}^{real}} \times 100\% \qquad (28)$$

where $d_{sub,i}^{sim}$ and $d_{sub,i}^{real}$ represent the lane-changing distance for the $i$-th lane change in the simulation and real-world data, respectively. $L_{sub}^{sim}$ and $L_{sub}^{real}$ denote the total number of lane changes observed in the simulation and real-world data.

(ii) Error in background vehicles' lane-changing distance: The relative error between simulated and real-world average lane-changing distance of the background vehicles. It is calculated based on the sampled data from surrounding vehicles. It is defined as follows:

$$d_{bg}^{sim} = \frac{1}{N_{surr}^{sim}} \sum_{k=1}^{N_{surr}^{sim}} \frac{\sum_{i=1}^{L_k^{sim}} d_{k,i}^{sim}}{L_k^{sim}} \qquad (29)$$

$$d_{bg}^{real} = \frac{1}{N_{surr}^{real}} \sum_{k=1}^{N_{surr}^{real}} \frac{\sum_{t=1}^{L_k^{real}} d_{k,t}^{real}}{L_k^{real}} \qquad (30)$$

$$e_{d,bg} = \frac{|d_{bg}^{sim} - d_{bg}^{real}|}{d_{bg}^{real}} \times 100\% \qquad (31)$$

where $d_{bg}^{sim}$ is the simulated average lane-changing distance of background vehicles, $d_{k,i}^{sim}$ is the lane-changing distance of the $i$-th lane change by the $k$-th surrounding vehicle in the simulation, $L_k^{sim}$ denotes the number of lane changes performed by that vehicle, and $N_{surr}^{sim}$ is the number of surrounding vehicles with observable lane changes in the simulation. $d_{bg}^{real}$ is the corresponding value from real-world data, using analogous notation. $e_{d,bg}$ is the relative error.

(iii) Frequency of cut-ins: The relative error between simulated and real-world frequency of cut-ins experienced by the subject vehicle. It is defined by:

$$e_{cutin} = \begin{cases} 0, \Delta n \leq \Delta_{LB} \\ \dfrac{\Delta n - \Delta_{LB}}{\Delta_{UB} - \Delta_{LB}}, & \Delta_{LB} < \Delta n < \Delta_{UB} \\ 100\%, \Delta n \geq \Delta_{UB} \end{cases} \quad (32)$$

where $\Delta n = |n_{cutin}^{real} - n_{cutin}^{sim}|$ represents the absolute difference in the frequency of cut-ins between real-world and simulation data. $\Delta_{LB}$ and $\Delta_{UB}$ are the piecewise points of $\Delta n$. In this study, $\Delta_{LB}$ is set to 0 and $\Delta_{UB}$ is set to 1.

To be noted, the frequency of cut-ins in Equation (32) adopted is a fuzzy membership function, as illustrated in Figure 6, to normalize the error within the range of [0,100%]. This ensures comparability with other MoEs.

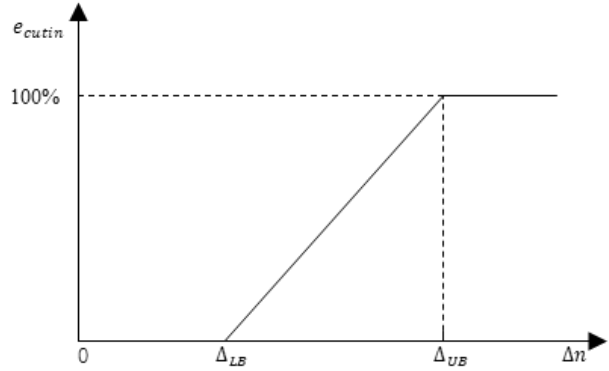

**Figure 6** The fuzzy membership function of $e_{cutin}$.

**Traffic flow-level accuracy**

The traffic flow-level accuracy measures the calibration capability of traffic flow. It adopts the following MoEs:

(i) Traffic speed: The relative error between simulated and real-world traffic speed, as defined by:

$$e_v = \frac{|v_{traffic}^{real} - v_{traffic}^{sim}|}{v_{traffic}^{real}} \times 100\% \quad (33)$$

where $v_{traffic}^{sim}$ and $v_{traffic}^{real}$ denote the traffic speed in the real-world data and simulation data, respectively. They are calculated based on the sampled data of the subject vehicle and its surrounding vehicles.

(ii) Traffic density: The relative error in traffic density, as defined by:

$$e_\rho = \frac{|\rho_{traffic}^{real} - \rho_{traffic}^{sim}|}{\rho_{traffic}^{real}} \times 100\% \quad (34)$$

where $\rho_{traffic}^{sim}$ and $\rho_{traffic}^{real}$ denote the density in the simulation and the real-world sampled data, respectively.

(iii) Traffic volume: The relative error in traffic volume, as defined by:

$$e_{vol} = \frac{|f_{traffic}^{real} - f_{traffic}^{sim}|}{f_{traffic}^{real}} \times 100\% \qquad (35)$$

where $f_{traffic}^{sim}$ and $f_{traffic}^{real}$ represent the traffic volume observed in the simulation and the real-world sampled data, respectively.

(iv) Traffic dynamic: The MoE evaluates the consistency of the spatial-temporal evolution of background traffic, under the condition that the subject vehicle follows the same origin-destination routes as in the real world. It is quantified by the relative error of the subject vehicle's average speed between the simulation and the real world. It is defined by:

$$e_{dynamics} = \frac{|v_{sub}^{real} - v_{sub}^{sim}|}{v_{sub}^{real}} \times 100\% \qquad (36)$$

where $v_{sub}^{real}$ and $v_{sub}^{sim}$ represent the average speeds of the subject vehicle in the real-world and simulation data, respectively.

**Calibration efficiency**

Calibration efficiency is quantified by calibration time (Paz et al., 2015). Calibration time is defined as the duration from the start of case generation in Stage 1 to the output of the optimal parameter values in Stage 2.

To evaluate the robustness of the proposed method in terms of calibration efficiency across various traffic patterns, a similarity index referred to (Bag et al., 2019) is defined by:

$$J_m(A, B) = \frac{|A \cap B|}{K} \qquad (37)$$

where $A$ and $B$ are two sets of identified calibration parameters under two different traffic patterns, and $K$ is the total number of parameters in a set. This similarity index ($J_m(A, B)$) quantifies the consistency of the identified calibration parameters under different traffic conditions. The similarity ranges from 0 to 1, with higher values indicating greater overlap. Since calibration efficiency is strongly influenced by both the number of parameters and their initial values, higher consistency across traffic patterns suggests that the calibration process is more likely to require similar time durations, thereby indicating the robustness of the proposed method.

## 5.4 Parameter settings

The parameter settings are as follows:
**Simulation platform settings:**
- Total simulation time: 1750 *sec*.
- Warm-up time: 600 *sec*.
- Detection range of the subject vehicle: [-150 *m*, 150 *m*].
- Time step: 0.1 *sec*.
- Desired speed of the subject vehicle: 60 *km/h*

**The proposed calibration method settings:**
- Orthogonal arrays: $L_{64}(4^{12})$ for Stage 1 and $L_{64}(4^{20})$ for Stage 2. To be noted, the

number of cases is 64, the levels of each parameter is 4, the number of parameters is 12 in Stage 1 and 20 in Stage 2.
- Scaling factor ($\delta$): 0.2
- Number of significant parameters ($K$): 10
- Population size in each generation ($N_p$): 20
- Number of generations ($G_{max}$): 15
- Adaptive crossover probability bounds ($P_{c,min}$, $P_{c,max}$): [0.6, 0.9]
- Adaptive mutation probability bounds ($P_{m,min}$, $P_{m,max}$): [0.05, 0.25]
- Accuracy threshold ($a_{thre}$): 0.8

## 5.5 Results

The experimental results confirm that the proposed method successfully meets the outlined objectives: accurately replicating vehicle-to-vehicle interactions, ensuring accuracy at both the vehicle and traffic flow levels, and enhancing calibration efficiency. Specifically: i) The proposed method improves vehicle-level accuracy by 83.53% compared to the SOTA method, demonstrating its capability of calibration for vehicle-to-vehicle interactions. ii) It enhances accuracy at both the traffic flow level and vehicle level, with average improvements of 51.9% over the SOTA method, confirming its capability of assuring accuracy. iii) It reduces calibration time by 76.75%, confirming its capability of improving calibration efficiency. A detailed analysis of the results and some interesting findings are provided in the remainder of this section.

### 5.5.1 Traffic flow-level accuracy

Figure 7 illustrates the performance of the proposed method in calibrating traffic flow-level metrics. While it does not outperform all baselines across every metric, it achieves significant improvements in two key areas: traffic density (4.2% error reduction) and traffic dynamics (3.4% error reduction). These were intentionally prioritized due to their relevance to AV evaluation—density influences the likelihood of interaction-heavy scenarios such as lane changes and cut-ins, while dynamics ensure temporal realism in the traffic conditions experienced by the AV. The method shows slightly lower accuracy in traffic volume and speed, which were deliberately excluded from the calibration objectives. This trade-off is intentional: volume and speed represent macroscopic traffic properties and overlook the individual-level behaviors crucial to AV performance. Calibrating all four metrics simultaneously would weaken the focus on interaction fidelity. Therefore, the modest reduction in volume and speed accuracy is acceptable to achieve higher precision in density and dynamics. Despite this targeted focus, the method delivers overall performance on par with existing approaches, demonstrating its effectiveness and practical value for AV-oriented traffic flow calibration.

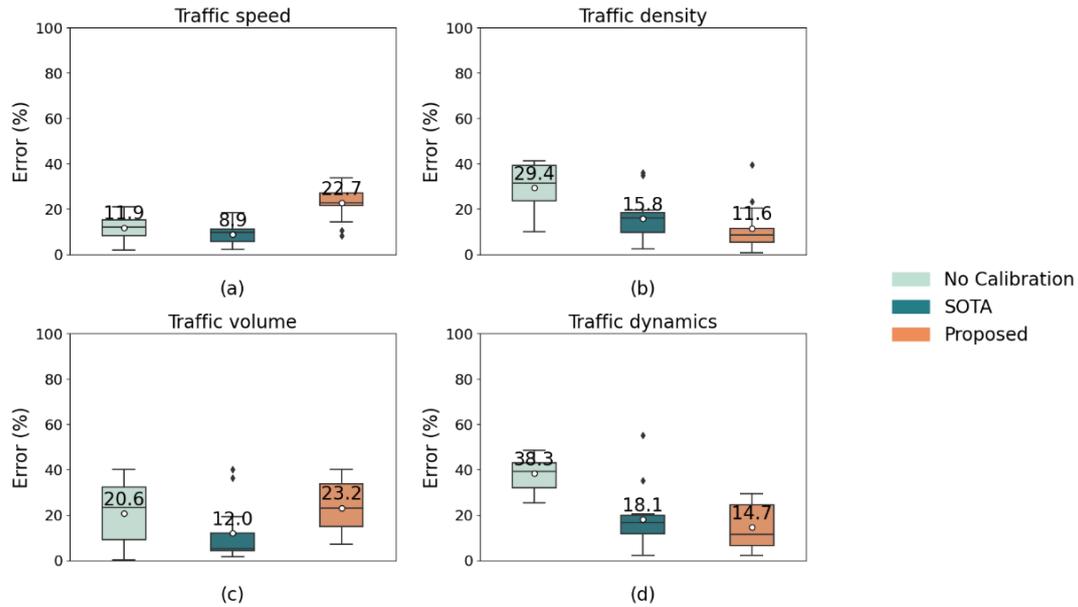

Figure 7 Performance on traffic flow-level accuracy.

### 5.5.2 Vehicle-level accuracy

Figure 8 presents the performance on vehicle-level accuracy. It demonstrates the proposed method's capability of calibration for vehicle-to-vehicle interactions. As shown, the proposed method reduces the total error by 85.22% compared to the No Calibration, and by 83.53% relative to the SOTA method. It confirms the effectiveness of incorporating vehicle-level MoPs into the calibration objectives.

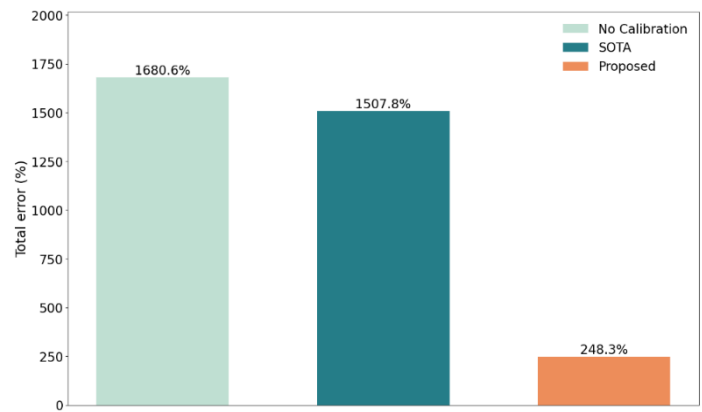

Figure 8 Performance on vehicle-level accuracy.

Figure 9 provides a detailed breakdown of vehicle-level accuracy in both longitudinal and lateral aspects. Figure 9(a) and (b) present the longitudinal performance. The results show that the proposed method does not exhibit statistically significant improvement over other methods in longitudinal accuracy. Figure 9(c)-(e) present the lateral performance. The proposed method achieves statistically significant improvements in lateral accuracy, reducing errors by 74.2% to 402.4% compared to No Calibration, and by 69.2% to 327.9% compared to the SOTA method.

A notable disparity is observed between the improvements in longitudinal and lateral aspects, with significantly greater enhancements in lateral behavior replication. This discrepancy arises from the differing nature of factors influencing longitudinal and lateral interactions, leading to varying sensitivities of vehicle-level Measures of Performance (MoPs) to calibration. Longitudinal interactions—such as headway—are primarily shaped by traffic flow characteristics. Under comparable traffic flow conditions (as shown in Figure 7), longitudinal behaviors tend to be more homogeneous across vehicles, resulting in limited variability. Consequently, their accuracy exhibits lower sensitivity to calibration on vehicle-level MoPs. In contrast, lateral interactions—such as lane changes and cut-ins—are influenced not only by traffic flow but also by vehicle-specific factors, including driver intent and preferences. These factors vary widely across vehicles, making lateral interactions more diverse and thus more sensitive to vehicle-level calibration. Therefore, by explicitly incorporating vehicle-level MoPs into the calibration, the proposed method achieves greater improvement in lateral accuracy. This highlights the critical role of vehicle-level calibration in lateral accuracy.

A more comprehensive analysis covering both traffic flow-level accuracy and vehicle-level accuracy is provided in Section 5.5.3.

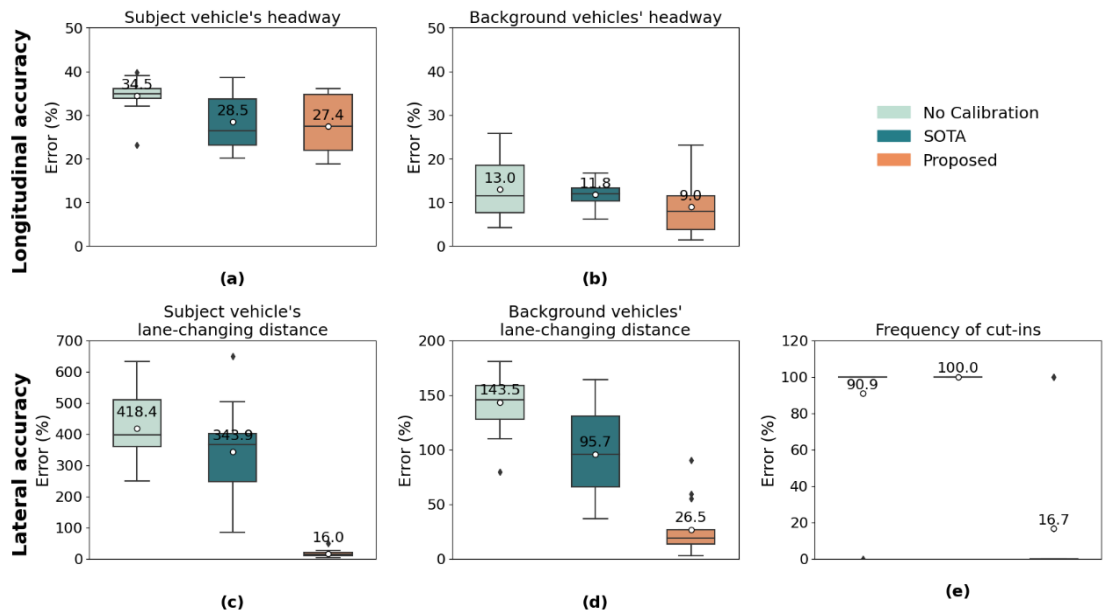

**Figure 9** Performance on longitudinal accuracy and lateral accuracy.

### 5.5.3 Discussion on accuracy performance

Figure 10 summarizes the accuracy improvements achieved by the proposed method across both vehicle-level and traffic flow-level accuracy. It demonstrates the capability to assure accuracy on both levels. As shown in Figure 9(a), the proposed method improves accuracy by an average of 70.3% compared to No Calibration. Figure 9(b) shows an average improvement of 51.9% over the SOTA method.

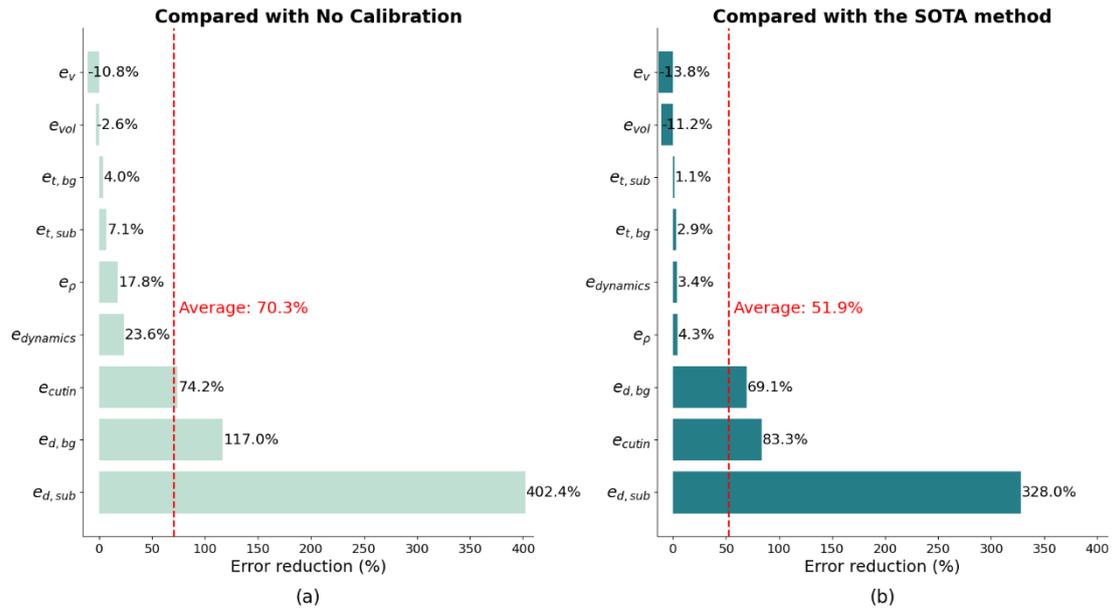

**Figure 10** Accuracy achievement achieved by the proposed method.

However, a trade-off is observed: the improvement in vehicle-level accuracy comes at the cost of a slight reduction in traffic flow-level accuracy, with two traffic flow-level MoEs exhibiting degradation. This trade-off stems from the calibration decomposition strategy employed by the proposed method, which consists of two stages: Stage 1 focuses on traffic flow-level accuracy, while Stage 2 targets vehicle-level accuracy. During Stage 2, when vehicle behavior parameters are calibrated, traffic flow characteristics can be influenced. For instance, increasing the desired following distance may improve vehicle-level accuracy but lower traffic volume, thereby impairing the traffic flow-level accuracy. As a result, the calibration in Stage 2 may disrupt the traffic flow-level balance established in Stage 1, resulting in degradations in traffic flow-level accuracy.

Despite this trade-off, the proposed method achieves a balanced accuracy performance. As shown in Figure 11, the overall error across all evaluation aspects ranges from 9.0% to 27.4%. This demonstrates that the proposed method is practical for simulation tests requiring both traffic flow-level and vehicle-level accuracy.

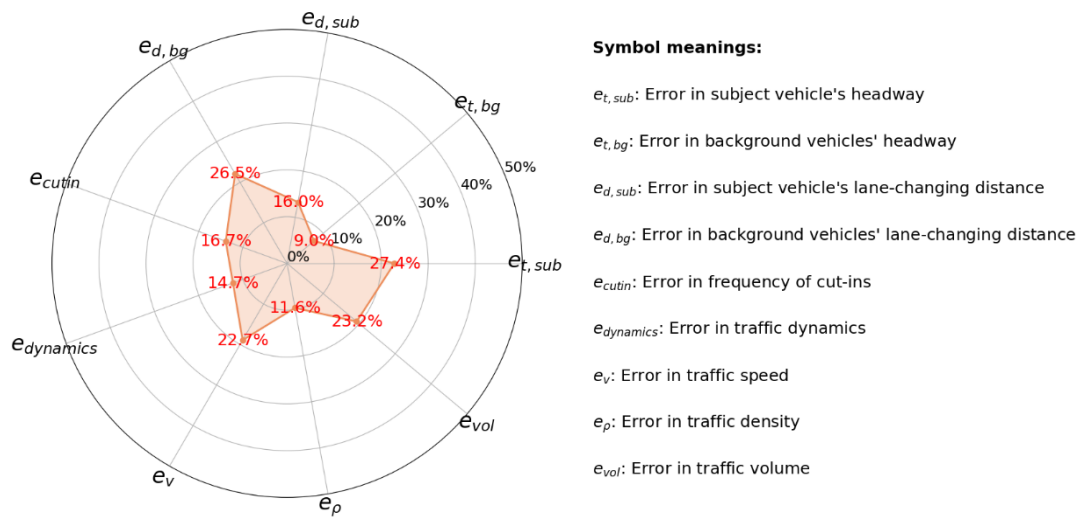

**Figure 11** Performance on accuracy

### 5.5.4 Calibration efficiency

Figure 12 illustrates the performance on calibration efficiency. The results highlight the proposed method's capability of enhancing efficiency. The proposed method improves efficiency by 76.75% compared to the state-of-the-art method. This improvement is attributed to the incorporation of two strategies: calibration decomposition and only critical cases are tested. These strategies reduce the overall number of cases that need to be tested, thereby enhancing efficiency.

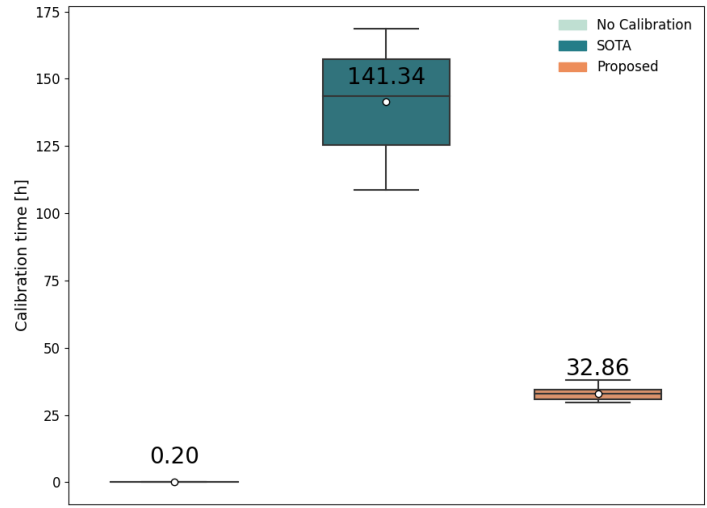

**Figure 12** Performance on calibration efficiency.

One interesting finding is that the proposed method's calibration efficiency is robust to variations in traffic patterns. As shown in Figure 12, the proposed method exhibits 2.38 hours of standard deviation in calibration time while the state-of-the-art method exhibits 19.12 hours. This robustness is attributed to the proposed method's capability to consistently identify a similar set of calibration parameters, regardless of traffic patterns.

Figure 13 presents the heatmap of the similarity index, which quantifies the consistency of identified calibration parameters under different traffic patterns. As shown in the figure, 100% of traffic pattern pairs share at least three calibration parameters (i.e., all similarity values are ≥ 0.3), and over 80% share at least half of the identified calibration parameters (similarity values ≥ 0.5). This high similarity results in similar solution spaces across traffic patterns, as the structure of the solution space is determined by the selected parameters and their respective ranges. Given these similar solution spaces, the proposed method follows comparable search trajectories and converges in a similar number of iterations, thereby leading to consistent calibration times.

In contrast, the state-of-the-art method lacks a parameter selection design. It randomly selects parameters for optimization without evaluating their importance to calibration accuracy, resulting in different parameter sets under varying traffic patterns. Consequently, the optimization is carried out in highly inconsistent solution spaces, leading to unstable convergence behavior and large variations in calibration time. This limits the method's robustness in calibration efficiency.

In summary, the proposed method not only improves calibration efficiency but also ensures robustness against traffic pattern variations. These advantages make it well-suited for large-scale or time-sensitive simulation tests where both efficiency and robustness are essential.

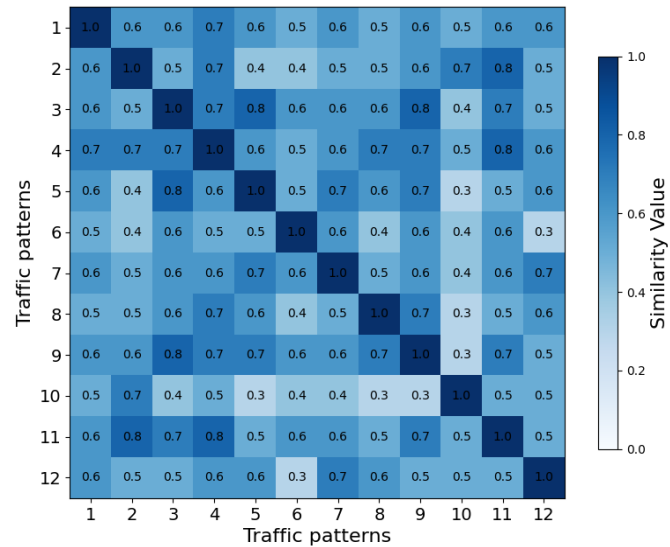

**Figure 13** The modified Jaccard similarity heatmap across various traffic patterns.

# 6 Conclusion

This research proposes a simulation platform calibration method that enables accuracy on vehicle level and traffic flow level. The proposed method has the following features: i) with the capability of calibration for vehicle-to-vehicle interaction; ii) with accuracy assurance; iii) with efficiency consideration; iv) with pipeline calibration capability. A comprehensive evaluation was conducted by simulations. The proposed method has been compared against the commonly used calibration methods. Results demonstrated that:

- The proposed method improves vehicle-level accuracy by 96.88% compared to the state-of-the-art method, validating its capability of calibration for vehicle-to-vehicle interactions.
- The proposed method achieves accuracy improvements of 51.9% over the state-of-the-art method, confirming its capability to ensure calibration accuracy.
- The proposed method improves calibration efficiency by 76.75% compared to the state-of-the-art method.
- The proposed method's calibration efficiency is robust to varying traffic patterns.
- The proposed method enables fully automated calibration, eliminating the need for human intervention.
- A notable disparity is observed between the improvements in longitudinal and lateral replication, with significantly greater enhancements in lateral behavior replication.
- A trade-off is observed: the improvement in vehicle-level accuracy comes at the cost of a slight reduction in traffic flow-level accuracy, with two traffic flow-level MoEs exhibiting degradation.

# Acknowledgements

This paper is partially supported by National Key R&D Program of China (2022YFF0604905), National Natural Science Foundation of China (Grant No. 52372317), Yangtze River Delta Science and Technology Innovation Joint Force (No. 2023CSJGG0800), Shanghai Automotive Industry Science and Technology Development Foundation (No. 2404), Xiaomi Young Talents Program, the Fundamental Research Funds for the Central Universities (22120230311), Tongji Zhongte Chair Professor Foundation (No. 000000375-2018082), and the Science Fund of State Key Laboratory of Advanced Design and Manufacturing Technology for Vehicle (No. 32215011).

# Appendix A Suggested MoPs

Table 2 lists the suggested vehicle-level MoPs, offering references for accurately replicating vehicle-to-vehicle interactions.

**Table 2** Suggested vehicle-level MoPs.

| Interaction behavior | Vehicle object | Statistic quantity | Maneuver features | The suggested MoPs |
|---|---|---|---|---|
| Cut-in | The subject vehicle | Count, average value, distribution, median value, percentiles, etc. | The frequency of being cut-in, The distance when cut-ins occur | The number of cut-in, etc. |
| Overtaking | The subject vehicle, interactive vehicles | | The frequency of being overtaking, The distance when overtaking occurs | The number of being overtaking, etc. |
| Lane change | | | The frequency of lane-changing, The distance of lane-changing, The duration of lane-changing | The average lane-changing distance of the subject vehicle, etc. |
| Car-following | | | The frequency of car-following, The distance of car-following, The duration of car-following, The spacing of car-following, The speed of the following vehicle, The speed oscillation | The average car-following headway of subject vehicle, etc. |

Table 3 lists the suggested traffic-level MoPs, offering references for accurately replicating traffic flow encountered by the subject vehicle.

Table 3 Suggested traffic-level MoPs.

| Traffic characteristic | Range | Statistic quantity | The suggested evaluation |
|---|---|---|---|
| Flow (volume) | On the passed road(s) by subject vehicles (a single road or all roads) | Average value | The average flow on a specific road, etc. |
| Density | | | The average density across all the traveled roads, etc. |
| Average speed | | | The average speed across all the traveled roads, etc. |
| Travel time of the subject vehicle | | | The average speed of the subject vehicle across all the traveled roads, etc. |

# Appendix B Vehicle models and suggested calibration parameters

Table 4 lists the most used vehicle models and their suggested parameters, offering references for selection of available vehicle behavior parameters.

Table 4 Examples of vehicle models and their calibration parameters.

| Category | Name | Suggested parameters | Reference |
|---|---|---|---|
| Automated vehicle model | An integrated behavioral and motion planner with proactive right-of-way acquisition capability | $\theta_1$: The weight of driving safety<br>$\theta_2$: The weight of travel efficiency<br>$\theta_3$: The weight of ride comfort | (Zhang et al., 2023) |
| | Intelligent driver model (IDM) | $a_{max}$: The maximum acceleration<br>$b$: The comfortable deceleration<br>$T$: Desired time gap<br>$s_0$: The jam space gap<br>$\delta$: The acceleration exponent | (Treiber et al., 2000) |
| | Asymmetric optimal velocity model with a relative velocity term (AOVRV model) | $k_{1D}$, $k_{2D}$: The gains on spacing and speed difference when the speed difference is less than 0;<br>$k_{1A}$, $k_{2A}$: The gains on spacing and speed difference when the speed difference is larger than 0<br>$T$: Desired time gap | (Shang et al., 2022) |
| Conventional vehicle model | Gazi-Herman-Rothery (GHR) model | $\alpha$: constant sensitivity coefficient<br>$\beta$: sensitivity to following vehicle (FV) speed<br>$\gamma$: sensitivity to space headway | (Gazis et al., 1961) |

|  | | $\tau_n$: reaction time |  |
|  | Gipps model | $\tilde{a}_n$: maximum desired acceleration of FV<br>$\tilde{b}_n$: maximum desired deceleration of FV<br>$\delta_{n-1}$: Effective length of leading vehicle (LV)<br>$\hat{b}$: maximum desired deceleration of LV<br>$\tilde{V}_n$: desired speed of FV<br>$\tau_n$: reaction time | (Peter G. Gipps, 1981) |
|  | Full velocity difference model (FVD) | $\alpha$: Constant sensitivity coefficient<br>$\lambda_0$: Sensitivity to relative speed<br>$V_0$: Desired speed pf FV<br>$b$: Interaction length<br>$\beta$: Form factor<br>$S_c$: Max following distance | (Jiang et al., 2001) |
|  | Wiedemann 99 model | $CC0$: standstill gap<br>$CC1$: Headway time<br>$CC2$: Following variation<br>$CC3$: Threshold for entering following<br>$CC4$: Negative following threshold<br>$CC5$: Positive following threshold<br>$CC6$: Speed dependency of oscillation<br>$CC7$: Oscillation acceleration<br>$CC8$: Standstill acceleration<br>$CC9$: Acceleration at 80km/h | (Durrani et al., 2016) |
|  | VISSIM lane-changing model | SafeDistRedFact: Safety distance reduction factor<br>MinHdwy: Minimum headway (front/rear)<br>Max.DecelTrail: Maximum deceleration for trailing vehicle<br>Max.DecelOwn: Maximum deceleration for own vehicle | (Al-Msari et al., 2024) |
|  | Krauss model | $a$: The driver's preferred maximum acceleration<br>$b$: The maximum deceleration<br>$\tau$: The reaction time | (J. Song et al., 2014) |

# References

Alhariqi, A., Gu, Z., & Saberi, M. (2022). Calibration of the intelligent driver model (IDM) with adaptive parameters for mixed autonomy traffic using experimental trajectory data. Transportmetrica B: Transport Dynamics, 10(1), 421–440. https://doi.org/10.1080/21680566.2021.2007813

Ali, Y., Zheng, Z., & Bliemer, M. C. J. (2023). Calibrating lane-changing models: Two data-related issues and a general method to extract appropriate data. Transportation Research Part C: Emerging Technologies, 152, 104182. https://doi.org/10.1016/j.trc.2023.104182

Al-Msari, H., Koting, S., Ahmed, A. N., & El-shafie, A. (2024). Review of driving-behaviour simulation: VISSIM and artificial intelligence approach. Heliyon, 10(4), e25936. https://doi.org/10.1016/j.heliyon.2024.e25936

Bag, S., Kumar, S. K., & Tiwari, M. K. (2019). An efficient recommendation generation using relevant Jaccard similarity. Information Sciences, 483, 53–64. https://doi.org/10.1016/j.ins.2019.01.023

Chao Zhang, Carolina Osorio, & Gunnar Flötteröd. (2017). Efficient calibration techniques for large-scale traffic simulators. Transportation Research Part B: Methodological, 97, 214–239. https://doi.org/10.1016/j.trb.2016.12.005

Chiappone, S., Giuffrè, O., Granà, A., Mauro, R., & Sferlazza, A. (2016). Traffic simulation models calibration using speed–density relationship: An automated procedure based on genetic algorithm. Expert Systems with Applications, 44, 147–155. https://doi.org/10.1016/j.eswa.2015.09.024

Dai, Q., Shen, D., Wang, J., Huang, S., & Filev, D. (2022). Calibration of human driving behavior and preference using vehicle trajectory data. Transportation Research Part C: Emerging Technologies, 145, 103916. https://doi.org/10.1016/j.trc.2022.103916

Ding, W., Zhang, L., Chen, J., & Shen, S. (2022). EPSILON: An Efficient Planning System for Automated Vehicles in Highly Interactive Environments. IEEE Transactions on Robotics, 38(2), 1118–1138. IEEE Transactions on Robotics. https://doi.org/10.1109/TRO.2021.3104254

Durrani, U., Lee, C., & Maoh, H. (2016). Calibrating the Wiedemann's vehicle-following model using mixed vehicle-pair interactions. Transportation Research Part C: Emerging Technologies, 67, 227–242. https://doi.org/10.1016/j.trc.2016.02.012

Gazis, D. C., Herman, R., & Rothery, R. W. (1961). Nonlinear Follow-the-Leader Models of Traffic Flow. Operations Research, 9(4), 545–567. https://doi.org/10.1287/opre.9.4.545

Guthrie, W. F. (2020). NIST/SEMATECH e-Handbook of Statistical Methods (NIST Handbook 151) [Dataset]. National Institute of Standards and Technology. https://doi.org/10.18434/M32189

Hale, D. K., Ghiasi, A., Khalighi, F., Zhao, D., Li, X. (Shaw), & James, R. M. (2023). Vehicle Trajectory-Based Calibration Procedure for Microsimulation. Transportation Research Record, 2677(1), 1764–1781. https://doi.org/10.1177/03611981221124597

Han, N. er lan M., Qin, Y., Zhang, Q. hua, Yang, Y. fang, & Zhang, Z. dong. (2013). Parameter Calibration of VISSIM Simulation Model Based on Genetic Algorithm. 591–596. https://doi.org/10.2991/icacsei.2013.142

Jiang, R., Wu, Q., & Zhu, Z. (2001). Full velocity difference model for a car-following theory. Physical Review E, 64(1), 017101. https://doi.org/10.1103/PhysRevE.64.017101

Kesting, A., & Treiber, M. (2008). Calibrating Car-Following Models by Using Trajectory Data: Methodological Study. Transportation Research Record: Journal of the Transportation Research Board, 2088(1), 148–156. https://doi.org/10.3141/2088-16

Langer, M., Harth, M., Preitschaft, L., Kates, R., & Bogenberger, K. (2021). Calibration and Assessment of Urban Microscopic Traffic Simulation as an Environment for Testing of Automated Driving. 2021 IEEE International Intelligent Transportation Systems Conference (ITSC), 3210–3216. https://doi.org/10.1109/ITSC48978.2021.9564743

Li, H. X., Li, Y., Jiang, B., Zhang, L., Wu, X., & Lin, J. (2020). Energy performance optimisation of building envelope retrofit through integrated orthogonal arrays with data envelopment analysis.

Renewable Energy, 149, 1414–1423. https://doi.org/10.1016/j.renene.2019.10.143

Li, L., Chen, X. (Micheal), & Zhang, L. (2016). A global optimization algorithm for trajectory data based car-following model calibration. Transportation Research Part C: Emerging Technologies, 68, 311–332. https://doi.org/10.1016/j.trc.2016.04.011

Mohamed, H. A., Puan, O. C., Hassan, S. A., & Azhari, S. F. (2021). Calibration and Validation of Lane Changing Following Gap Distance in VISSIM. IOP Conference Series: Materials Science and Engineering, 1153(1), 012016. https://doi.org/10.1088/1757-899X/1153/1/012016

Paz, A., Molano, V., Martinez, E., Gaviria, C., & Arteaga, C. (2015). Calibration of traffic flow models using a memetic algorithm. Transportation Research Part C: Emerging Technologies, 55, 432–443. https://doi.org/10.1016/j.trc.2015.03.001

Peter G. Gipps. (1981). A behavioural car-following model for computer simulation—ScienceDirect. https://www.sciencedirect.com/science/article/abs/pii/0191261581900370

Pourabdollah, M., Bjärkvik, E., Fürer, F., Lindenberg, B., & Burgdorf, K. (2017). Calibration and evaluation of car following models using real-world driving data. 2017 IEEE 20th International Conference on Intelligent Transportation Systems (ITSC), 1–6. https://doi.org/10.1109/ITSC.2017.8317836

Punzo, V., & Montanino, M. (2020). A two-level probabilistic approach for validation of stochastic traffic simulations: Impact of drivers' heterogeneity models. Transportation Research Part C: Emerging Technologies, 121, 102843. https://doi.org/10.1016/j.trc.2020.102843

Punzo, V., & Simonelli, F. (2005). Analysis and Comparison of Microscopic Traffic Flow Models with Real Traffic Microscopic Data.

Ren, W., Zhao, X., Li, H., & Fu, Q. (2024). Traffic flow impact of mixed heterogeneous platoons on highways: An approach combining driving simulation and microscopic traffic simulation. Physica A: Statistical Mechanics and Its Applications, 643, 129803. https://doi.org/10.1016/j.physa.2024.129803

Sadid, H., & Antoniou, C. (2024). A simulation-based impact assessment of autonomous vehicles in urban networks. IET Intelligent Transport Systems, 18(9), 1677–1696. https://doi.org/10.1049/itr2.12537

Shang, M., Rosenblad, B., & Stern, R. (2022). A Novel Asymmetric Car Following Model for Driver-Assist Enabled Vehicle Dynamics. IEEE Transactions on Intelligent Transportation Systems, 23(9), 15696–15706. IEEE Transactions on Intelligent Transportation Systems. https://doi.org/10.1109/TITS.2022.3145292

Shikun, L., Yi, T., & Yonghong, L. (2024). Application of Driving Simulation Technology in Calibration of Traffic Simulation Parameters. Journal of System Simulation, 36(6), 1359. https://doi.org/10.16182/j.issn1004731x.joss.23-1549

Song, G., Yu, L., & Xu, L. (2013). Comparative Analysis of Car-Following Models for Emissions Estimation. Transportation Research Record: Journal of the Transportation Research Board, 2341(1), 12–22. https://doi.org/10.3141/2341-02

Song, J., Wu, Y., Xu, Z., & Lin, X. (2014). Research on car-following model based on SUMO. The 7th IEEE/International Conference on Advanced Infocomm Technology, 47–55. https://doi.org/10.1109/ICAIT.2014.7019528

Tang, L., Zhang, D., Han, Y., Fu, A., Zhang, H., Tian, Y., Yue, L., Wang, D., & Sun, J. (2024). Parallel-Computing-Based Calibration for Microscopic Traffic Simulation Model. Transportation Research Record: Journal of the Transportation Research Board, 2678(4), 279–294.

https://doi.org/10.1177/03611981231184244

The New York Times. (2024). Waymo Turns Toward Making Money as It Expands Driverless Car Business—The New York Times. https://www.nytimes.com/2024/09/04/technology/waymo-expansion-alphabet.html

Treiber, M., Hennecke, A., & Helbing, D. (2000). Congested traffic states in empirical observations and microscopic simulations. Physical Review E - Statistical Physics, Plasmas, Fluids, and Related Interdisciplinary Topics, 62(2), 1805–1824. Scopus. https://doi.org/10.1103/PhysRevE.62.1805

U.S. DOT. (2021). Trajectory Investigation for Enhanced Calibration of Microsimulation Models. https://doi.org/10.21949/1521658

Waymo Driver. (2025). Self-Driving Car Technology for a Reliable Ride—Waymo Driver. Waymo. https://waymo.com/waymo-driver/

Wu, Q. H., Cao, Y. J., & Wen, J. Y. (1998). Optimal reactive power dispatch using an adaptive genetic algorithm. International Journal of Electrical Power & Energy Systems, 20(8), 563–569. https://doi.org/10.1016/S0142-0615(98)00016-7

Ye, L., & Yamamoto, T. (2018). Modeling connected and autonomous vehicles in heterogeneous traffic flow. Physica A: Statistical Mechanics and Its Applications, 490, 269–277. https://doi.org/10.1016/j.physa.2017.08.015

Yu, H., Chung, C. Y., & Wong, K. P. (2011). Robust Transmission Network Expansion Planning Method With Taguchi's Orthogonal Array Testing. IEEE Transactions on Power Systems, 26(3), 1573–1580. IEEE Transactions on Power Systems. https://doi.org/10.1109/TPWRS.2010.2082576

Zhang, Z., Yan, X., Wang, H., Ding, C., Xiong, L., & Hu, J. (2023). No more road bullying: An integrated behavioral and motion planner with proactive right-of-way acquisition capability. Transportation Research Part C: Emerging Technologies, 156, 104363. https://doi.org/10.1016/j.trc.2023.104363